\pdfoutput=1

\documentclass[11pt]{article}

\usepackage{emnlp2021}

\usepackage{times}
\usepackage{latexsym}
\usepackage{booktabs}
\usepackage{graphicx}
\usepackage{tikz}
\usepackage{pgfplots}
\usepackage{subcaption}
\usepackage[ruled, linesnumbered, vlined]{algorithm2e}
\usepackage{hyperref}
\SetKwInput{KwInput}{Input}                
\SetKwInput{KwOutput}{Output}
\usepackage{amsmath, amssymb}
\usepackage{xcite}
\usepackage{xr}
\makeatletter
\newcommand*{\addFileDependency}[1]{
  \typeout{(#1)}
  \@addtofilelist{#1}
  \IfFileExists{#1}{}{\typeout{No file #1.}}
}
\makeatother
\newcommand*{\myexternaldocument}[1]{%
    \externaldocument{#1}%
    \addFileDependency{#1.tex}%
    \addFileDependency{#1.aux}%
}

\myexternaldocument{Appendix}

\usepackage[T1]{fontenc}

\usepackage[utf8]{inputenc}

\usepackage{microtype}

\title{Learning Neural Ordinary Equations for Forecasting Future Links on Temporal Knowledge Graphs}

\author{Zhen Han\thanks{\; Equal contribution.}$\;^{1,2}$, \; Zifeng Ding$^{*1,2}$, \; Yunpu Ma$^{*1}$,\; Yujia Gu$^{3}$ \; Volker Tresp\thanks{\; Corresponding author.}$\;^{ 1,2}$ \\
$^{1}$Institute of Informatics, LMU Munich $\;$  $^{2}$ Corporate Technology, Siemens AG\\
$^{3}$Department of Electrical and Computer Engineering, Technical University of Munich\\
\texttt{zhen.han@campus.lmu.de,  cognitive.yunpu@gmail.com}\\
\texttt{\{zifeng.ding, volker.tresp\}@siemens.com, yujia.gu@tum.de}\\}

\begin{document}
\maketitle

\begin{abstract}
There has been an increasing interest in inferring future links on temporal knowledge graphs (KG). While links on temporal KGs vary continuously over time, the existing approaches model the temporal KGs in discrete state spaces. To this end, we propose a novel continuum model by extending the idea of neural ordinary differential equations (ODEs) to multi-relational graph convolutional networks. The proposed model preserves the continuous nature of dynamic multi-relational graph data and encodes both temporal and structural information into continuous-time dynamic embeddings. In addition, a novel graph transition layer is applied to capture the transitions on the dynamic graph, i.e., edge formation and dissolution. We perform extensive experiments on five benchmark datasets for temporal KG reasoning, showing our model's superior performance on the future link forecasting task.
\end{abstract}

\section{Introduction}
Reasoning on relational data has long been considered an essential subject in artificial intelligence with wide applications, including decision support and question answering. Recently, reasoning on knowledge graphs has gained increasing interest \cite{ren2020beta, das2018walk}. A Knowledge Graph (KG) is a graph-structured knowledge base to store factual information. KGs represent facts in the form of triples $(s,r,o)$, e.g., (\textit{Bob}, \textit{livesIn}, \textit{New York}), in which $s$ (subject) and $o$ (object) denote nodes (entities), and $r$ denotes the edge type (relation) between $s$ and $o$. Knowledge graphs are commonly static and store facts in their current state. In reality, however, the relations between entities often change over time. For example, if Bob moves to California, the triple of (\textit{Bob}, \textit{livesIn}, \textit{New York}) will be invalid. To this end, temporal knowledge graphs (tKG) were introduced. A tKG represents a temporal fact as a quadruple $(s,r,o,t)$ by extending a static triple with time $t$, describing that this fact is valid at time $t$. In recent years, several sizable temporal knowledge graphs, such as ICEWS \cite{DVN/28075_2015}, 
have been developed that provide widespread availability of such data and enable reasoning on temporal KGs. While lots of work  \cite{garcia-duran-etal-2018-learning, goel2020diachronic, lacroix2020tensor} focus on the temporal KG completion task and predict missing links at observed timestamps, recent work \cite{jin2019recurrent, trivedi2017knowevolve} paid attention to forecast future links of temporal KGs. In this work, we focus on the temporal KG forecasting task, which is more challenging than the completion task.

Most existing work \cite{jin2019recurrent, zhu2020learning} models temporal KGs in a discrete-time domain where they take snapshots of temporal KGs sampled at regularly-spaced timestamps. Thus, these approaches cannot model irregular time intervals, which convey essential information for analyzing dynamics on temporal KGs, e.g., the dwelling time of a user on a website becomes shorter, indicating that the user's interest in the website decreases.  KnowEvolve \cite{trivedi2017knowevolve} uses a neural point process to model continuous-time temporal KGs. However, Know-Evolve does not take the graph's structural information into account, thus losing the power of modeling temporal topological information. Also, KnowEolve is a transductive method that cannot handle unseen nodes. In this paper, we present a graph neural-based approach to learn dynamic representations of entities and relations on temporal KGs. Specifically, we propose a graph neural ordinary differential equation to model the graph dynamics in the continuous-time domain. 

Inspired by neural ordinary differential equations (NODEs) \cite{chen2018neural}, we extend the idea of continuum-depth models to encode the continuous dynamics of temporal KGs. 
To apply NODEs to temporal KG reasoning, we employ a NODE coupled with multi-relational graph convolutional (MGCN) layers. MGCN layers are used to capture the structural information of multi-relational graph data, while the NODE learns the evolution of temporal KGs over time. Specifically, we integrate the hidden representations over time using an ODE solver and output the continuous-time dynamic representations of entities and relations. 
Unlike many existing temporal KG models that learn the dynamics by employing recurrent model structures with discrete depth, our model lets the time domain coincide with the depth of a neural network and takes advantage of NODE to steer the latent entity features between two timestamps smoothly. 
Besides,  existing work simply uses the adjacency tensor from previous snapshots of the tKG to predict its linkage structure at a future time. Usually, most edges do not change between two observations, while only a few new edges have formatted or dissolved since the last observation. However, the dissolution and formation of these small amounts of edges always contain valuable temporal information and are more critical than unchanged edges for learning the graph dynamics. For example, we know an edge with the label \textit{economicallyCooperateWith} between two countries $x$ and $y$ at time $t$, but this dissolves at $t+\Delta t_1$. Additionally, there is another edge with the label \textit{banTradesWith} between these two countries that are formated at $t+\Delta t_2$ ($\Delta t_2>\Delta t_1$). Intuitively, the dissolution of ($x$, \textit{economicallyCooperateWith}, $y$) is an essential indicator of the quadruple ($x$, \textit{banTradesWith}, $y$, $t+\Delta t_2$). Thus, it should get more attention from the model. 
However, suppose we only feed the adjacency tensors of different observation snapshots into the model. In that case, we do not know whether the model can effectively capture the changes of the adjacency tensors and puts more attention on the evolving part of the graph. To let the model focus on the graph's transitions, we propose a graph transition layer that takes a graph transition tensor containing edge formation and dissolution information as input and uses graph convolutions to process the transition information explicitly.

In this work, we propose a model to perform \textbf{T}emporal Knowledge Gr\textbf{a}ph Forecasti\textbf{ng} with Neural \textbf{O}rdinary Equations (TANGO\includegraphics[height=0.8em]{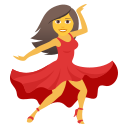}). The main contributions are summarized as follows:
\begin{itemize}
\item We propose a continuous-depth multi-relational graph neural network for forecasting future links on temporal KGs by defining a multi-relational graph neural ordinary differential equation. The ODE enables our model to learn continuous-time representations of entities and relations. We are the first to show that the neural ODE framework can be extended to modeling dynamic multi-relational graphs.
\item We propose a graph transition layer to model the edge formation and dissolution of temporal KGs, which effectively improves our model's performance.
\item We propose two new tasks, i.e., inductive link prediction and long horizontal link forecasting, for temporal KG models. They evaluate a model's potential by testing the model's performance on previously unseen entities and predicting the links happening in the farther future.
\item We apply our model to forecast future links on five benchmark temporal knowledge graph datasets, showing its state-of-the-art performance.
\end{itemize}

\section{Preliminaries and Related Work} 
\subsection{Graph Convolutional Networks}
Graph convolutional networks (GCNs) have shown great success in capturing structural dependencies of graph data. GCNs come in two classes: $i)$ spectral methods \citep{kipf2016semi, defferrard2016convolutional} and $ii)$ spatial methods \citep{niepert2016learning, gilmer2017neural}.  
 However, common GCNs can only deal with homogeneous graphs. To distinguish between different relations, R-GCN \cite{schlichtkrull2017modeling} introduces relation-specific weight matrices for message transformations. However, the number of parameters in R-GCN grows rapidly with the number of relations, easily leading to overfitting. \citet{vashishth2019composition} proposed a multi-relational GCN, which is compatible with KGs and leverages various entity-relation composition operations from KG embedding techniques. Additionally, some work combines GCN with temporal graphs \citep{yan2018spatial, li2020temporal}. However, they are designed for homogeneous graphs but not for multi-relational graphs.

\subsection{Neural Ordinary Differential Equations}
Neural Ordinary Differential Equation (NODE) \cite{chen2018neural} is a continuous-depth deep neural network model. It represents the derivative of the hidden state with a neural network:
\begin{equation}
    \frac{d\textbf{z}(t)}{d t} = f(\textbf{z}(t), t, \theta),
\end{equation}
where $\textbf{z}(t)$ denotes the hidden state of a dynamic system at time $t$, and $f$ denotes a function parameterized by a neural network to describe the derivative of the hidden state regarding time. $\theta$ represents the parameters in the neural network. The output of a NODE framework is calculated using an ODE solver coupled with an initial value:
\begin{equation}
    \textbf{z}(t_1) = \textbf{z}(t_0) + \int_{t_0}^{t_1}f(\textbf{z}(t), t, \theta)dt.
\end{equation}
Here, $t_0$ is the initial time point, and $t_1$ is the output time point. $\textbf{z}(t_1)$ and $\textbf{z}(t_0)$ represent the hidden state at $t_1$ and $t_0$, respectively. Thus, the NODE can output the hidden state of a dynamic system at any time point and deal with continuous-time data, which is extremely useful in modeling continuous-time dynamic systems.

Moreover, to reduce the memory cost in the backpropagation, \citet{chen2018neural} introduced the adjoint sensitivity method into NODEs. An adjoint is $\textbf{a}(t) = \frac{\partial\mathcal{L}}{\partial \textbf{z}(t)}$, where $\mathcal{L}$ means the loss. The gradient of $\mathcal{L}$ with regard to network parameters $\theta$ can be directly computed by the adjoint and an ODE solver:
\begin{equation}
    \frac{d\mathcal{L}}{d\theta} = - \int_{t_1}^{t_0} \textbf{a}(t)^T\frac{\partial f(\textbf{z}(t),t,\theta)}{\partial \theta}dt.
\end{equation}
In other words, the adjoint sensitivity method solves an augmented ODE backward in time and computes the gradients without backpropagating through the operations of the solver. 

\subsection{Temporal Knowledge Graph Reasoning}
Let $\mathcal V$ and $\mathcal R$ represent a finite set of entities and relations, respectively. A temporal knowledge graph (tKG) $\mathcal G$ is a multi-relational graph whose edges evolve over time.  At any time point, a snapshot $\mathcal G(t)$ contains all valid edges at $t$.  Note that the time interval between neighboring snapshots may not be regularly spaced. 
A quadruple $q = (s, r, o, t)$ describes a labeled timestamped edge at time $t$, where $r \in \mathcal R$ represents the relation between a subject entity $s \in \mathcal V$ and an object entity $o \in \mathcal V$.  Formally, we define the tKG forecasting task as follows. 
Let $(s_q, r_q, o_q, t_q)$ denote a target quadruple and $\mathcal F$ represent the set of all ground-truth quadruples. Given query $(s_q, r_q, ?, t_q) $ derived from the target quadruple and a set of observed events $\mathcal O = \{(s, r, o, t_i) \in \mathcal F| t_i < t_q\}$, the tKG forecasting task predicts the missing object entity $o_q$ based on observed \textbf{past} events. Specifically, we consider all entities in set $\mathcal V$ as candidates and rank them by their scores to form a true quadruple together with the given subject-relation-pair ($s_q, r_q$) at time $t_q$. In this work, we add reciprocal relations for every quadruple, i.e., adding $(o,r^{-1}, s,t)$ for every $(s,r,o,t)$. Hence, the restriction to predict object entities does not lead to a loss of generality.

Extensive studies have been done for temporal KG \textbf{completion} task \citep{leblay2018deriving, garcia-duran-etal-2018-learning, goel2020diachronic, han2020dyernie}.
Besides, a line of work \cite{trivedi2017knowevolve, jin2019recurrent, deng2020dynamic, zhu2020learning} has been proposed for the tKG \textbf{forecasting} task and can generalize to unseen timestamps.
Specifically, \citet{trivedi2017knowevolve} and \citet{han2020graph} take advantage of temporal point processes to model the temporal KG as event sequences and learn evolving entity representations. 

\section{Our Model}
\begin{figure}
    \centering
    \begin{subfigure}[b]{0.5\textwidth}
    \centering
    \includegraphics[width=\textwidth]{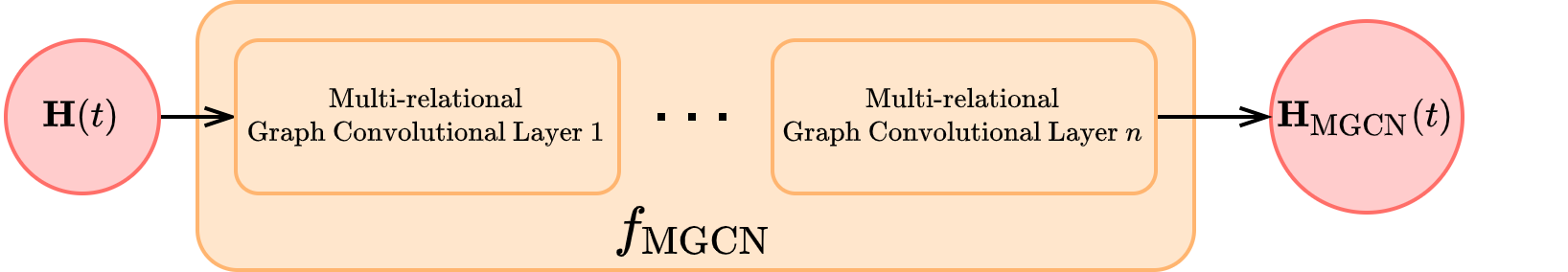}
    \caption{$f_{\textrm{MGCN}}$}
    \label{fig:F_MGCN}
    \end{subfigure}\\
    \vspace{2mm}
    \begin{subfigure}[b]{0.5
    \textwidth}
    \centering
    \includegraphics[width=.8\textwidth]{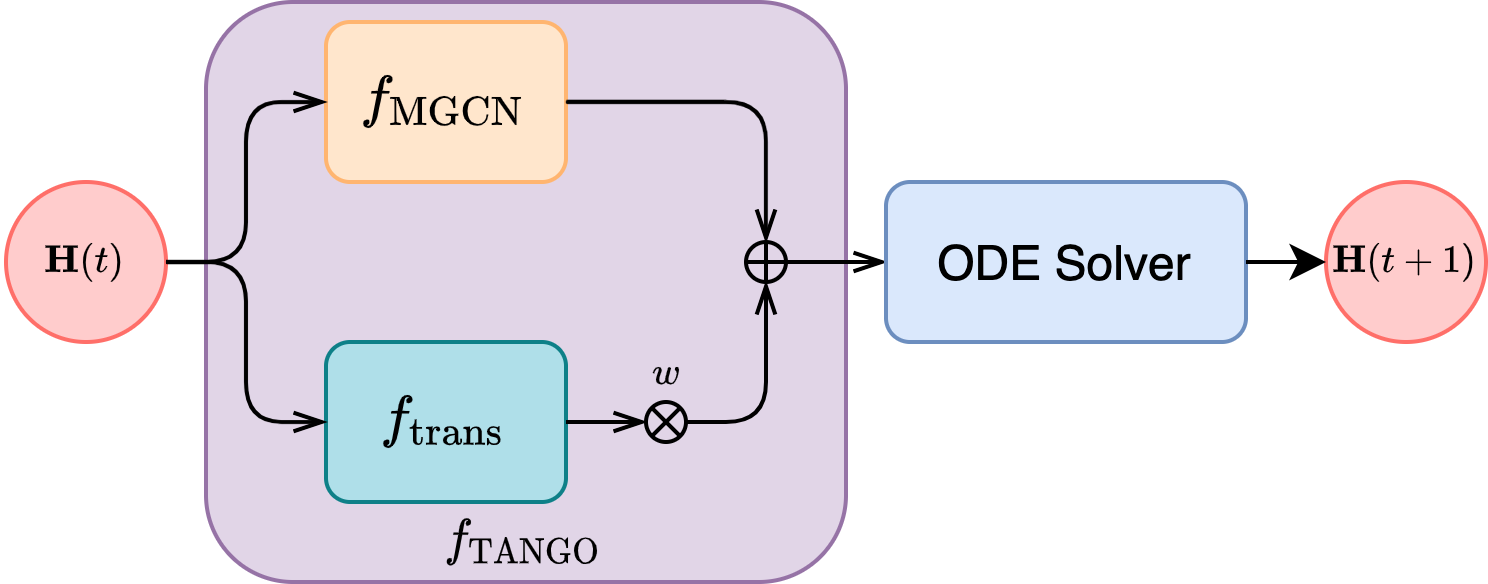}
    \caption{$f_{\textrm{TANGO}}$}
     \label{fig:network}
    \end{subfigure}
    \caption{(a) The structure of $f_{\textrm{MGCN}}$: stacked multi-relational graph convolutional layers (the orange block). $\textbf{H}(t)$ denotes the hidden representations of entities and relations at time $t$. $\textbf{H}_{\textrm{MGCN}}(t)$ denotes the output of the stacked multi-relational graph convolutional layers. 
    (b) The architecture of TANGO that parameterizes the derivatives of the hidden representations $\mathbf H(t)$. In addition to $f_{\textrm{MGCN}}$, a graph transition layer $f_{\textrm{trans}}$ is employed to model the edge formation and dissolution. 
}
\end{figure}

Our model is designed to model time-evolving multi-relational graph data by learning continuous-time representations of entities. It consists of a neural ODE-based encoder and a decoder based on classic KG score functions. As shown in Figure \ref{fig:network}, the input of the network will be fed into two parallel modules before entering the ODE Solver. The upper module denotes a multi-relational graph convolutional layer that captures the graph's structural information according to an observation at time $t$. And the lower module denotes a graph transition layer that explicitly takes the edge transition tensor of the current observation representing which edges have been added and removed since the last observation. The graph transition layer focuses on modeling the graph \textit{transition} between neighboring observations for improving the prediction of link formation and dissolution. For the decoder, we compare two score functions, i.e., DistMult \cite{yang2014embedding} and TuckER \cite{balazevic-etal-2019-tucker}. In principle, the decoder can be any score function. 

\subsection{Neural ODE for Temporal KG}
The temporal dynamics of a time-evolving multi-relational graph can be characterized by the following neural ordinary differential equation
\begin{equation}
\begin{aligned}
\label{equa: ODE}
    \frac{d\textbf{H}(t)}{dt} = &f_{\textrm{TANGO}}(\textbf{H}(t), \textbf{T}(t), \mathcal G(t), t) \\
    =&f_{\textrm{MGCN}}(\textbf{H}(t), \mathcal G(t), t) \\
    &+ w f_{\textrm{trans}}(\textbf{H}(t), \textbf{T}(t), \mathcal G(t), t), 
    \end{aligned}
\end{equation} 
where $\textbf{H} \in R^{(|\mathcal V| + 2|\mathcal R|)\times d}$ denotes the hidden representations of entities and relations. 
$f_{\textrm{TANGO}}$ represents the neural network that parameterizes the derivatives of the hidden representations. Besides, $f_{\textrm{MGCN}}$ denotes stacked multi-relational graph convolutional layers, $f_{\textrm{trans}}$ represents the graph transition layer, and $\mathcal G (t)$ denotes the snapshot of the temporal KG at time $t$. $\textbf{T}(t)$ contains the information on edge formation and dissolution since the last observation. $w$ is a hyperparameter controlling how much the model learns from edge formation and dissolution. We set $ \textbf{H}(t = 0) = \textrm{Emb}(\mathcal V, \mathcal R)$, where Emb($\mathcal V, \mathcal R$) denotes the learnable initial embeddings of entities and relations on the temporal KG. Thus, given a time window $\Delta t$, the representation evolution performed by the neural ODE assumes the following form
\begin{equation}
\begin{aligned}
&\textbf{H}(t+\Delta t) - \textbf{H}(t) \\ 
&= \int_t^{t+\Delta t}  f_{\textrm{TANGO}}(\textbf{H}(\tau), \textbf{T}(\tau), \mathcal G(\tau), \tau) \,d\tau\\
&= \int_t^{t+\Delta t} ( f_{\textrm{MGCN}}(\textbf{H}(\tau), \mathcal G(\tau), \tau) \\
& \quad \quad \quad \quad  \ \  +w f_{\textrm{trans}}(\textbf{H}(\tau), \textbf{T}(\tau), \tau)) d\tau. 
\end{aligned}
\end{equation}
In this way, we use the neural ODE to learn the dynamics of continuous-time temporal KGs. 

\subsection{Multi-Relational Graph Convolutional Layer}
\label{sec: multi-relation graph layer}
 Inspired by \cite{vashishth2019composition} and \cite{yang2014embedding}, we  use the entity-relation composition to model relational information. Specifically, we propose a multi-relational graph convolutional layer as follows. At time $t$, for every object entity $o\in \mathcal{V}$ with $\mathcal{N}(o) = \{(s,r)|(s,r,o,t)\in \mathcal{G}(t)\}$, its hidden representation evolves as 
\begin{equation}
\label{equa: entity convolution}
    \begin{aligned}
        &\widetilde{\textbf{h}}_o^{l+1}(t) = \frac{1}{|\mathcal{N}(o)|} \sum \limits_{(s,r) \in \mathcal{N}(o)} \textbf{W}^l (\textbf{h}_s^l (t) * \textbf{h}_r), \\
        &\textbf{h}_o^{l+1}(t) = \mathbf h_o^l(t) + \delta \sigma ( \widetilde{\textbf{h}}_o^{l+1}(t)),\\
    \end{aligned}
\end{equation}
where $\mathbf h_o^{l+1} (t)$ denotes the hidden representation of the object $o$ at the $(l+1)^{th}$ layer, $\textbf{W}^l$ represents the weight matrix on the $l^{th}$ layer, $*$ denotes element-wise multiplication. $\mathbf h_s^{l} (t)$ means the hidden representation of the subject $s$ at the $l^{th}$ layer.
$\mathbf h^{l=0}_s(t) = \mathbf h_s(t)$ is obtained by the ODE Solver that integrates Equation \ref{equa: ODE} until $t$.  $\delta$ is a learnable weight.
In this work, we assume that the relation representations do not evolve, and thus, $\mathbf h_r$ is time-invariant. We use $ ReLU(\cdot)$ as the activation function $\sigma(\cdot)$. 
From the view of the whole tKG, we use $\mathbf H(t)$ to represent the hidden representations of all entities and relations on the tKG. Besides, we use $f_{\textrm{MGCN}}$ to denote the network consisting of multiple multi-relational graph convolutional layers (Equation \ref{equa: entity convolution}). 

\subsection{Graph Transition Layer}
\label{sec: jump layer}
To let the model focus on the graph's transitions, we define a transition tensor for tKGs and use graph convolutions to capture the information of edge formation and dissolution. Given two graph snapshots $\mathcal G(t-\Delta t)$ and $\mathcal G(t)$ at time $t - \Delta t$ and $t$, respectively, the graph transition tensor $\mathbf{T}(t)$ is defined as  
\begin{equation}
    \textbf{T}(t) = \textbf{A}(t) -  \textbf{A}(t-\Delta t),
\end{equation}
where $\textbf{A}(t) \in \{0,1\}^{|\mathcal{V}|\times |\mathcal R| \times |\mathcal{V}|}$ is a three-way adjacency tensor whose entries are set such that
\begin{equation}
A_{sro} = 
\left \{
\begin{aligned}
    &1, \textrm{if the triple $(s, r, o)$ exists at time $t$},\\
    &0, \textrm{otherwise}.
\end{aligned}
\right .
\end{equation}
Intuitively, $\textbf{T}(t) \in \{-1,0,1\}^{|\mathcal{V}|\times |\mathcal R| \times |\mathcal{V}|}$ contains the information of the edges' formation and dissolution since the last observation $\mathcal G(t - \Delta t)$. Specifically, $T_{sro} (t) = -1$ means that the triple $(s, r, o)$ disappears at $t$, and $T_{sro}(t) = 1$ means that the triplet $(s, r, o)$ is formatted at $t$.  
For all unchanged edges, their values in $\textbf{T}(t)$ are equal to $0$. 
Additionally, we use graph convolutions to extract the information provided by the graph transition tensor:
\begin{equation}
\begin{aligned}
     &\widetilde{\textbf{h}}^{l+1}_{o, \textrm{trans}} (t) =  \textbf{W}_{\textrm{trans}} (T_{sro}(t)(\textbf{h}^l_s(t) * \textbf{h}_r)) \\
    &\textbf{h}^{l+1}_{o, \textrm{trans}} (t) = \sigma \left(\frac{1}{|\mathcal{N}_T(o)|}\sum \limits_{(s,r) \in \mathcal{N}_T(o)} \widetilde{\textbf{h}}^{l+1}_{o, \textrm{trans}} (t)  \right ) 
\end{aligned}
\label{equa: trans layer}
\end{equation}
Here, $\textbf{W}_{\textrm{trans}}$ is a trainable diagonal weight matrix and $\mathcal{N}_{T}(o) = \{(s,r)|T_{sro}(t)\neq 0)\}$. By employing this graph transition layer, we can better model the dynamics of temporal KGs. We use $f_{\textrm{trans}}$ to denote Equation \ref{equa: trans layer}. By combining the multi-relational graph convolutional layers $f_{\textrm{MGCN}}$ with the graph transition layer $f_{\textrm{trans}}$, we get our final network that parameterizes the derivatives of the hidden representations $\textbf{H}(t)$, as shown in Figure \ref{fig:network}.

\subsection{Learning and Inference}
TANGO \includegraphics[height=0.8em]{TANGO.png} is an autoregressive model that forecasts the entity representation at time $t$ by utilizing the graph information before $t$. To answer a link forecasting query $(s, r, ?, t)$, TANGO takes three steps. First, TANGO computes the hidden representations $\mathbf H(t)$ of entities and relations at the time $t$. Then TANGO uses a score function to compute the scores of all quadruples $\{(s, r, o, t)|o \in \mathcal V \}$ accompanied with candidate entities. Finally, TANGO chooses the object with the highest score as its prediction. 

\paragraph{Representation inference} The representation inference procedure is done by an ODE Solver, which is $\mathbf H(t) = \textrm{ODESolver}(\textbf{H}(t-\Delta t), f_{\textrm{TANGO}}, t - \Delta t, t, \Theta_{\textrm{TANGO}}, \mathcal G)$. Adaptive ODE solvers may incur massive time consumption in our work. To keep the training time tractable, we use fixed-grid ODE solvers coupled with the Interpolated Reverse Dynamic Method (IRDM) proposed by \citet{daulbaev2020interpolated}. IRDM uses Barycentric Lagrange interpolation \cite{berrut2004barycentric} on Chebyshev grid \cite{tyrtyshnikov2012brief} to approximate the solution of the hidden states in the reverse-mode of NODE. Thus, IRDM can lower the time cost in the backpropagation and maintain good learning accuracy. Additional information about representation inference is provided in Appendix \ref{app: representation inference}.

\paragraph{Score function} Given the entity and relation representations at the query time $t_q$, one can compute the scores of every triple at $t_q$. In our work, we take two popular knowledge graph embedding models, i.e., Distmult \cite{yang2014embedding} and TuckER \cite{balazevic-etal-2019-tucker}. Given triple $(s,r,o)$, its score is computed as shown in Table \ref{tab: score function}. 

\begin{table}
\centering
\caption{Score Functions. $\textbf{h}_s, \textbf{h}_r, \textbf{h}_o$ denote the entity representations of the subject entity $s$, object entity $o$, and the representation of the relation $r$, respectively. $d$ denotes the hidden dimension of representations. $\mathcal{W} \in \mathbb{R}^{d \times d \times d}$ is the core tensor specified in \protect\cite{balazevic-etal-2019-tucker}
. As defined in \protect\cite{Tucker1964TheEO}, $\times_1, \times_2, \times_3$ are three operators indicating the tensor product in three different modes.}\label{tab: score function}
\resizebox{.48\textwidth}{!}{
\begin{tabular}{c c c} \hline
Method&Score Function& \\ \hline
Distmult \cite{yang2014embedding} & $<\textbf{h}_s, \textbf{h}_r, \textbf{h}_o>$ & $\textbf{h}_s, \textbf{h}_r, \textbf{h}_o \in \mathbb{R}^{d}$\\ \hline
TuckER \cite{balazevic-etal-2019-tucker} & $\mathcal{W}\times_1 \textbf{h}_s \times_2 \textbf{h}_r \times_3 \textbf{h}_o$ & $\textbf{h}_s, \textbf{h}_r, \textbf{h}_o \in \mathbb{R}^{d}$\\ \hline
\hline
\end{tabular}}
\end{table}

\paragraph{Parameter Learning} For parameter learning, we employ the cross-entropy loss:
\begin{equation}
    \mathcal{L} =  \sum \limits_{(s, r, o, t)\in \mathcal{F}} 
    -\textrm{log}(f(o|s,r,t,\mathcal V)),
\end{equation} 
where $f(o|s,r, t, \mathcal V) = \frac{\textrm{exp}(score(\mathbf h_s (t), \mathbf h_r, \mathbf h_o (t)))}{\sum \limits_{e\in \mathcal V} \textrm{exp}(score(\mathbf h_s (t), \mathbf h_r, \mathbf h_e (t)))}$. 
$e \in \mathcal{V}$ represents an object candidate, and $score(\cdot)$ is the score function. $\mathcal F$ summarizes valid quadruples of the given tKG. 

\section{Experiments}
\subsection{Experimental Setup}
We evaluate our model by performing future link prediction on five tKG datasets\footnote{Code and datasets are available at https://github.com/TemporalKGTeam/TANGO.}. We compare TANGO's performance with several existing methods and evaluate its potential with inductive link prediction and long horizontal link forecasting. Besides, an ablation study is conducted to show the effectiveness of our graph transition layer.
\subsubsection{Datasets}
\label{sec: datasets}
We use five benchmark datasets to evaluate TANGO: 1) ICEWS14 \cite{trivedi2017knowevolve} 2) ICEWS18 \cite{DVN/28075_2015} 3) ICEWS05-15 \cite{garcia-duran-etal-2018-learning} 4) YAGO \cite{mahdisoltani2013yago3} 5) WIKI \cite{leblay2018deriving}. 
Integrated Crisis Early Warning System (ICEWS) \cite{DVN/28075_2015} is a dataset consisting of timestamped political events, e.g., (\textit{Barack Obama}, \textit{visit}, \textit{India}, 2015-01-25). 
Specifically, ICEWS14 contains events occurring in 2014, while ICEWS18 contains events from January 1, 2018, to October 31, 2018. ICEWS05-15 is a long-term dataset that contains the events between 2005 and 2015. WIKI and YAGO are two subsets extracted from Wikipedia and YAGO3 \cite{mahdisoltani2013yago3}, respectively. 
The details of each dataset and the dataset split strategy are provided in Appendix \ref{app: dataset statistics}. 

\subsubsection{Evaluation Metrics}
\label{sec: evaluation metrics}
We use two metrics to evaluate the model performance on extrapolated link prediction, namely Mean Reciprocal Rank (MRR) and Hits@1/3/10. MRR is the mean of the reciprocal values of the actual missing entities' ranks averaged by all the queries, while Hits@1/3/10 denotes the proportion of the actual missing entities ranked within the top 1/3/10. The filtering settings have been implemented differently by various authors. We report results based on two common implementations: $i)$ time-aware \citep{han2021xerte} and $ii)$ time-unaware filtering \citep{jin2019recurrent}. We provide a detailed evaluation protocol in Appendix \ref{app: evaluation metrics}.

\subsubsection{Baseline Methods}
We compare our model performance with nine baselines. 
We take three static KG models as the static baselines, including Distmult \cite{yang2014embedding}, TuckER \cite{balazevic-etal-2019-tucker}, and COMPGCN \cite{vashishth2019composition}. 
For tKG baselines, we report the performance of TTransE \citep{leblay2018deriving}, TA-Distmult \cite{garcia-duran-etal-2018-learning},
CyGNet \citep{zhu2020learning}, DE-SimplE \citep{goel2020diachronic}, TNTComplEx \citep{lacroix2020tensor}, and RE-Net \cite{jin2019recurrent}. We provide implementation details of baselines and TANGO in Appendix \ref{app: Implementation Details}. 
 
\subsection{Experimental Results}

\subsubsection{Time-aware filtered Results}
\begin{table*}[t]
    \centering
    \resizebox{\textwidth}{!}{
    \large\begin{tabular}{@{}|l|cccc|cccc|cccc|cccc|cccc|@{}}
\toprule
        Datasets & \multicolumn{4}{|c}{\textbf{ICEWS05-15 - aware filtered}} &  \multicolumn{4}{|c}{\textbf{ICEWS14 - aware filtered}} & \multicolumn{4}{|c}{\textbf{ICEWS18 - aware filtered}} & \multicolumn{4}{|c}{\textbf{WIKI - aware filtered}}& \multicolumn{4}{|c|}{\textbf{YAGO - aware filtered}}\\
\midrule
        Model & MRR & Hits@1 & Hits@3 & Hits@10 & MRR & Hits@1 & Hits@3  & Hits@10 & MRR & Hits@1 & Hits@3  & Hits@10 & MRR & Hits@1 & Hits@3  & Hits@10 & MRR & Hits@1 & Hits@3  & Hits@10\\
\midrule 
        Distmult & 24.75 & 16.10 & 27.67 & 42.42 
        & 14.49 & 8.15 & 15.31 & 27.66 
        & 16.69 & 9.68 & 18.12 & 31.21
        & 49.66 & 46.17 & 52.81 & 54.13 
        & 54.84 & 47.39 & 59.81 & 68.52\\
        TuckER & 27.13 & 17.01 & 29.93 & 47.81 
        & 18.96 & 11.23 & 20.77 & 33.94 
        & 20.68 & 12.58 & 22.60 & 37.27
        & 50.01 & 46.12 & 53.60 & 54.86 
        & 54.86 & 47.42 & 59.63 & 68.96\\
        CompGCN & 29.68 & 20.72 & 32.51 & 47.87
        & 17.81 & 10.12 & 19.49 & 33.11 
        & 20.56 & 12.01 & 22.96 & 38.15
        & 49.88 & 45.78 & 52.91 & 55.58 
        & 54.35 & 46.72 & 59.26 & 68.29\\
\midrule
         TTransE & 21.24 & 4.98 & 31.48 & 49.88 
         & 9.67 & 1.25 & 12.29 & 28.37
         & 8.08 & 1.84 & 8.25 & 21.29 
         & 29.27 & 21.67 & 34.43 & 42.39 
         & 31.19 & 18.12 & 40.91 & 51.21\\
       TA-DistMult & 24.39 & 14.77 & 27.80 & 44.22 
       & 10.34 & 4.72 & 10.54 & 21.48 
       & 11.38 & 5.58 & 12.04 & 22.82 
       & 44.53 & 39.92 & 48.73 & 51.71 
       & 54.92 & 48.15 & 59.61 & 66.71\\
       
       CyGNet & 35.79 & 26.09 & 40.18 & 54.48 
       & 22.83 & 14.28 & 25.36 & 39.97 
       & 24.93 & 15.90 & 28.28 & 42.61
       & 33.89 & 29.06 & 36.10 & 41.86
       & 52.07 & 45.36 & 56.12 & 63.77\\
       
       DE-SimplE & 35.57 & 26.33 & 39.41 & 53.97
       & 21.58 & 13.77 & 23.68 & 37.15
       & 19.30 & 11.53 & 21.86 & 34.80 
       & 45.43 & 42.6 & 47.71 & 49.55 
       & 54.91 & 51.64 & 57.30 & 60.17 \\
       
       TNTComplEx & 35.88 & 26.92 & 39.55 & 53.43
       & 23.81 & 15.58 & 26.27 & 40.12
       & 21.23 & 13.28 & 24.02 & 36.91
       & 45.03 & 40.04 & 49.31 & 52.03 
       & 57.98 & 52.92 & 61.33 & 66.69 \\
       
       RE-Net & 40.23 & 30.30 & 44.83 & 59.59 
       & 25.66 & 16.69 & 28.35 & 43.62 
       & 27.90 & 18.45 & 31.37 & 46.37 
       & 49.66 & 46.88 & 51.19 & 53.48 
       & 58.02 & 53.06 & 61.08 & 66.29\\
\midrule
        \includegraphics[height=0.8em]{TANGO.png}-TuckER 
        & \textbf{42.86} & \textbf{32.72} & \textbf{48.14} & \textbf{62.34}
        & \textbf{26.25}  & \textbf{17.30} & \textbf{29.07} & \textbf{44.18} 
        & \textbf{28.97} & \textbf{19.51} & \textbf{32.61} & \textbf{47.51}  
        & 51.60 & 49.61 & 52.45 & 54.87 
        & 62.50 & 58.77 & 64.73 & 68.63\\
        &$\pm$ 0.2 & $\pm$ 0.3 & $\pm$ 0.2& $\pm$ 0.2
        &$\pm$ 0.1 & $\pm$ 0.1 & $\pm$ 0.1& $\pm$ 0.1
        &$\pm$ 0.2 & $\pm$ 0.1 & $\pm$ 0.2& $\pm$ 0.3
        &$\pm$ 0.3 & $\pm$ 0.2 & $\pm$ 0.3& $\pm$ 0.3
        &$\pm$ 0.5 & $\pm$ 0.2 & $\pm$ 0.1& $\pm$ 0.4\\
        \includegraphics[height=0.8em]{TANGO.png}-Distmult 
        & 40.71 & 31.23 & 45.33 & 58.95
        & 24.70  & 16.36 & 27.26 & 41.35
        & 27.56 & 18.68 & 30.86 & 44.94  
        & \textbf{53.04} & \textbf{51.52} & \textbf{53.84} & \textbf{55.46} 
        & \textbf{63.34} & \textbf{60.04} & \textbf{65.19} & \textbf{68.79}\\
        &$\pm$ 0.3 & $\pm$ 0.4 & $\pm$ 0.1& $\pm$ 0.5
        &$\pm$ 0.1 & $\pm$ 0.1 & $\pm$ 0.1& $\pm$ 0.1
        &$\pm$ 0.2 & $\pm$ 0.2 & $\pm$ 0.2& $\pm$ 0.3
        &$\pm$ 0.3 & $\pm$ 0.4 & $\pm$ 0.2& $\pm$ 0.1
        &$\pm$ 0.4 & $\pm$ 0.4 & $\pm$ 0.1& $\pm$ 0.2\\
\bottomrule
    \end{tabular}
    }
    \caption{Extrapolated link prediction results on five datasets. Evaluation metrics are time-aware
    filtered MRR (\%) and Hits@1/3/10 (\%). 
    \includegraphics[height=0.8em]{TANGO.png} denotes TANGO. 
    The best results are marked in bold. }\label{tab: td link prediction results}
\end{table*}
We run TANGO five times and report the averaged results. The time-aware filtered results are presented in Table \ref{tab: td link prediction results}, where \includegraphics[height=0.8em]{TANGO.png} denotes TANGO. As explained in Appendix \ref{app: evaluation metrics}, we take the time-aware filtered setting as the fairest evaluation setting.
Results demonstrate that TANGO \includegraphics[height=0.8em]{TANGO.png} outperforms all the static baselines on every dataset. This implies the importance of utilizing temporal information in tKG datasets. The comparison between Distmult and TANGO-Distmult shows the superiority of our NODE-based encoder, which can also be observed by the comparison between TuckER and TANGO-TuckER. Additionally, TANGO achieves much better results than COMPGCN, indicating our method's strength in incorporating temporal features into tKG representation learning.  

\begin{figure}[htbp]
\centering
\includegraphics[width=.7\columnwidth]{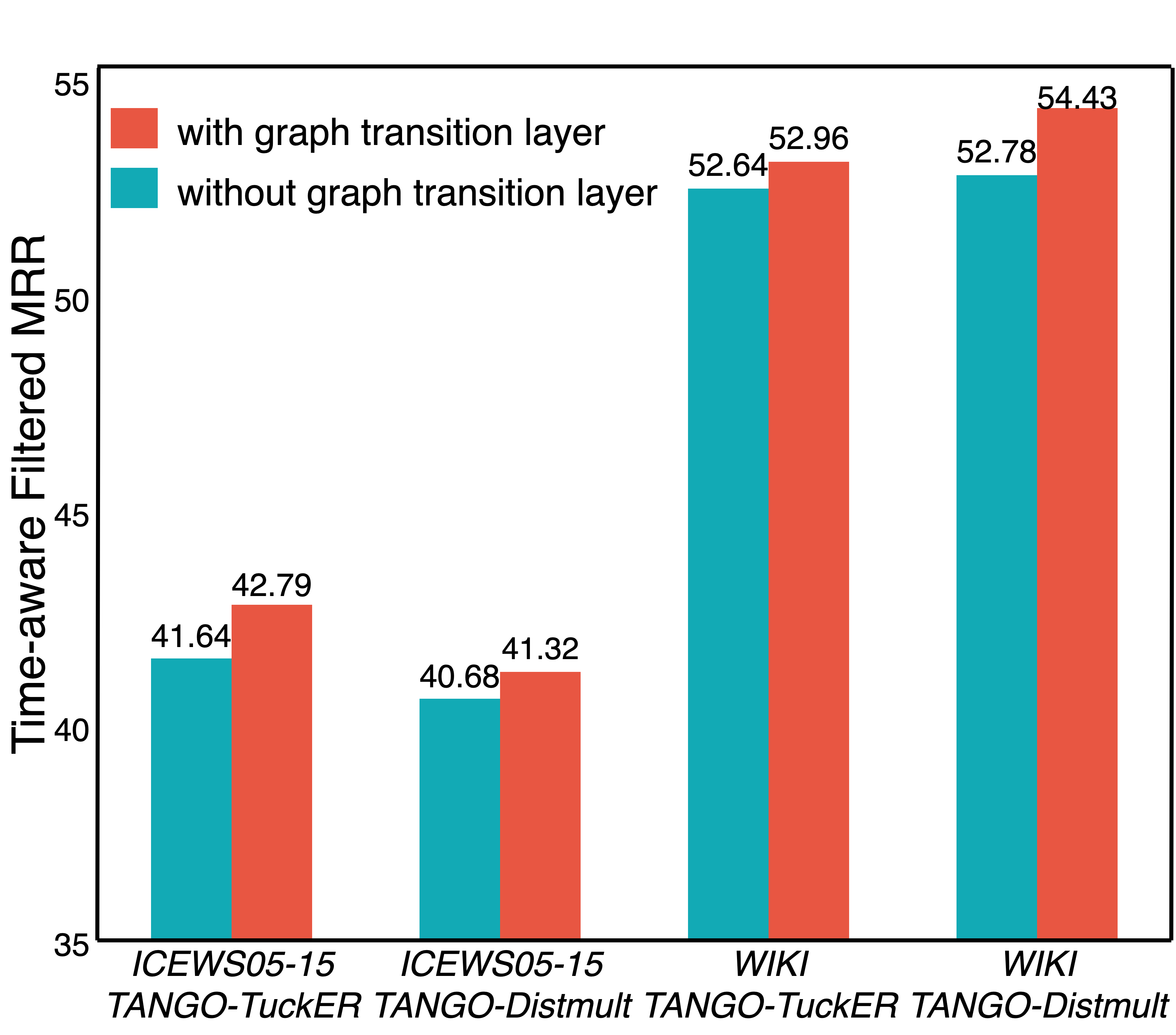}
\caption{Time-aware filtered MRR of TANGO with or without the graph transition layer on subsets of ICEWS05-15 and WIKI. We split the graph snapshots into two groups, where the transition tensor's norm $||\mathbf T(t)||_{L1}$ of each graph snapshot in the first group is larger than that of all graph snapshots in the second group. Since the graph transition layer is tailored to graph changes, we show the results of the first group here. The corresponding result of the ablation study on the whole test sets are presented in Figure \ref{fig: overall ablation.} in the appendix. }
\label{fig: ablation study}
\end{figure}

Similarly, TANGO outperforms all the tKG baselines as well. Unlike TTransE and TA-Distmult, RE-Net uses a recurrent neural encoder to capture temporal information, which shows great success on model performance and is the strongest baseline. Our model TANGO implements a NODE-based encoder in the recurrent style to capture temporal dependencies. It consistently outperforms RE-Net on all datasets because TANGO explicitly encodes time information into hidden representations while RE-Net only considers the temporal order between events. 
Additionally, we provide the raw and time-unaware filtered results in Table \ref{tab: tid-filtered link prediction results} and \ref{tab: raw link prediction results} in the appendix.



\subsubsection{Ablation Study}
To evaluate the effectiveness of our graph transition layer, we conduct an ablation study on two datasets, i.e., ICEWS05-15 and WIKI. We choose these two datasets as the representative of 
two types of tKG datasets.  ICEWS05-15 contains events that last shortly and happen multiple times, i.e., Obama visited Japan. In contrast, the events in the WIKI datasets last much longer and do not occur periodically, i.e., Eliran Danin played for Beitar Jerusalem FC between 2003 and 2010.  The improvement of the time-aware filtered MRR brought by the graph transition layer is illustrated in Figure \ref{fig: ablation study}, showing that the graph transition layer can effectively boost the model performance by incorporating the edge formation and dissolution information.

\begin{figure}[htbp]
\centering
\includegraphics[width=.6\columnwidth]{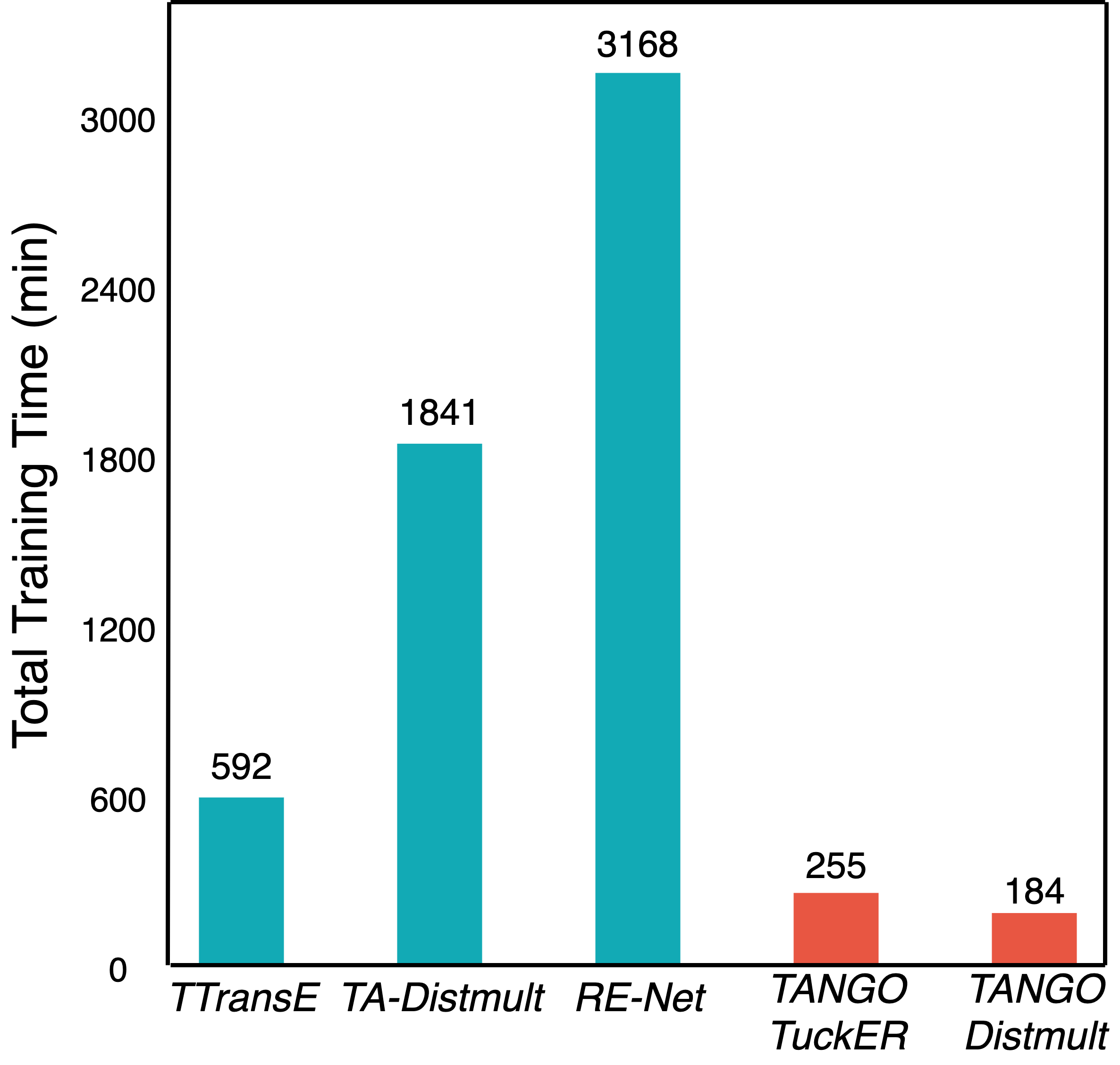}
\caption{\label{fig: time cost.}Time cost comparison on ICEWS05-15. Columns marked as orange denote the time consumed by our model.}
\end{figure}

\subsubsection{Time Cost Analysis} 
Keeping training time short while achieving a strong performance is significant in model evaluation. We report in Figure \ref{fig: time cost.} the total training time of our model and the baselines on ICEWS05-15. We see that static KG reasoning methods generally require less training time than temporal methods. Though the total training time for TTransE is short, its performance is low, as reported in the former sections. TA-Distmult consumes more time than our model and is also beaten by TANGO in performance. RE-Net is the strongest baseline in performance; however, it requires almost ten times as much as the total training time of TANGO. TANGO ensures a short training time while maintaining the state-of-the-art performance for future link prediction, which shows its superiority. 

\begin{table*}[htbp]
    \centering
    \resizebox{.9\textwidth}{!}{
    \large\begin{tabular}{@{}|l|cccc|cccc|cccc|@{}}
        \hline
        Datasets & \multicolumn{4}{|c}{\textbf{ICEWS05-15 - raw}}
        & \multicolumn{4}{|c}{\textbf{ICEWS05-15 - aware filtered}} 
        & \multicolumn{4}{|c|}{\textbf{ICEWS05-15 - unaware filtered}}\\
\toprule
        Model & MRR & Hits@1 & Hits@3 & Hits@10 & MRR & Hits@1 & Hits@3  & Hits@10 & MRR & Hits@1 & Hits@3  & Hits@10 \\
\midrule\relax
         RE-Net & 4.96 & 2.20 & 5.39 & 10.12 
         &  5.02 & 2.29 & 5.49 & 10.12 
         & 5.50 & 2.95 & 5.93 & 10.26 \\
       \includegraphics[height=0.8em]{TANGO.png}-TuckER w.o.trans & 5.13 & 2.58 & 5.67 & 9.91 
       & 5.18 &  2.64 &  5.70 & 9.94 
       & 5.98 & 3.34 & 6.71 & 10.67 
       \\
      \includegraphics[height=0.8em]{TANGO.png}-Distmult w.o.trans & 3.72 & 2.05 & 3.80 & 6.76 &
        3.76 & 2.09 & 3.82 & 6.77 &
        4.09 & 2.46 & 4.17 & 6.99 \\
      \includegraphics[height=0.8em]{TANGO.png}-TuckER & \textbf{5.74} & \textbf{3.07} & \textbf{6.48} & \textbf{10.74} &
        \textbf{5.81} & \textbf{3.16} & \textbf{6.52} & \textbf{10.78} &
        \textbf{6.75} & \textbf{4.11} & \textbf{7.60} & \textbf{11.54} \\
      \includegraphics[height=0.8em]{TANGO.png}-Distmult & 5.00 & 2.70 & 5.67 & 9.16 &
        5.05 & 2.78 & 5.69 & 9.17 &
        5.69 & 3.45 & 6.27 & 9.69 \\
\bottomrule
    \end{tabular}}
    \caption{Inductive future link prediction results on ICEWS05-15. Evaluation metrics are raw, time-aware filtered, and time-unaware filtered MRR (\%), Hits@1/3/10 (\%). 
    \textit{w.o.trans} means without the graph transition layer. The best results are marked in bold.}\label{tab: inductive link prediction results}
\end{table*}

\subsection{New Evaluation Tasks}
\subsubsection{Long Horizontal Link Forecasting}
Given a sequence of observed graph snapshots until time $t$, 
the future link prediction task infers the quadruples happening at $t+\Delta t$. $\Delta t$ is usually small, i.e., one day, in standard settings \citep{trivedi2017knowevolve, jin2019recurrent, zhu2020learning}. However, in some scenarios, the graph information right before the query time is likely missing. This arouses the interest in evaluating the temporal KG models by predicting the links in the farther future. In other words, given the same input, 
the model should predict the links happening at $t + \Delta T$, where $\Delta T >> \Delta t$. Based on this idea, we define a new evaluation task, e.g., long horizontal link forecasting. 

\begin{figure}[htbp]
\centering
\begin{tikzpicture}[scale=0.7, transform shape]
  \definecolor{clr1}{HTML}{12AAB5}
  \definecolor{clr2}{HTML}{E85642}
\begin{axis}[%
  xlabel=$\Delta T$ (Day),
  ylabel=Time-aware Filtered MRR,
  ymajorgrids=true,
  xtick={1,2,3,4,5,6,7,8},
  ytick={35,36,37,38,39,40,41,42,43},
  grid style=dashed,
  ymin=38.0, ymax=43.0,
  xmin=1.0,xmax=8.0, 
]
  \addplot+[mark=square*, mark options={fill=clr2}, clr2] coordinates {(1,42.86) (2,41.36) (3, 40.88) (4, 40.57) (5, 40.32) (6, 39.88) (7, 39.42) (8, 39.10)};
  \addlegendentry{TANGO-TuckER}
  \addplot+[mark=square*, mark options={solid, fill=clr1}, clr1] coordinates {(1,40.23) (2,39.76) (3,39.39) (4,39.13) (5, 39.00) (6, 38.85) (7, 38.65) (8,38.48)};
  \addlegendentry{RE-Net}
\end{axis}
\end{tikzpicture}
\caption{Long horizontal link forecasting: time-aware filtered MRR (\%) on ICEWS05-15 with regard to different $\Delta t$.}
\label{fig: long horizontal link forecasting.}
\end{figure}
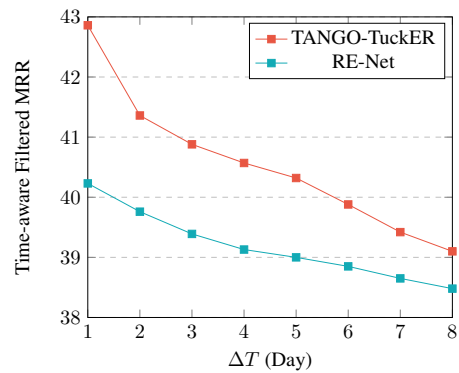

\begin{figure}[htbp]
    \centering
    \includegraphics[scale=0.5]{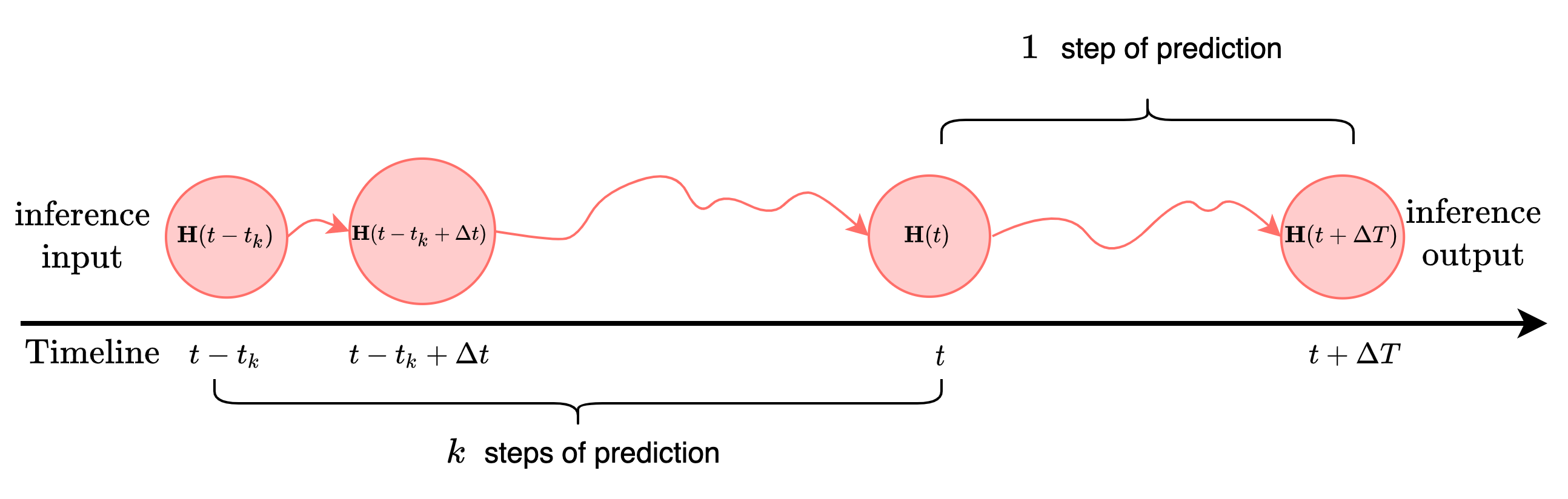}
    \caption{Graphical illustration of long horizontal link forecasting. Given a sequence of graph snapshots $\mathbb{G}=\{\mathcal{G}(t-t_k), ..., \mathcal{G}(t)\}$, whose length is $k$, test quadruples at $t+\Delta T$ are to be predicted.}
    \label{fig:Longinfer}
\end{figure}

To perform long horizontal link forecasting, we adjust the integral length according to how far the future we want to predict. As described in Figure \ref{fig:Longinfer}, 
the integration length between the neighboring timestamps is short for the first $k$ steps, e.g., integration from $(t-t_k)$ to $(t-t_k+\Delta t)$. However, for the last step, e.g., integration from $t$ to $t+\Delta T$, the integration length becomes significantly large according to how far the future we want to predict. The larger $\Delta T$ is, the longer the length is for the last integration step. 

We report the results corresponding to different $\Delta T$ on ICEWS05-15 and compare our model with the strongest baseline RE-Net. In Figure \ref{fig: long horizontal link forecasting.}, we observe that our model outperforms RE-Net in long horizontal link forecasting. The gap between the performances of the two models diminishes as $\Delta T$ increases. This trend can be explained in the following way. Our model employs an ODE solver to integrate the graph's hidden states over time. Since TANGO takes the time information into account and integrates the ODE in the continuous-time domain, its performance is better than RE-Net, which is a discrete-time model. However, TANGO assumes that the dynamics it learned at $t$ also holds at $t + \Delta T$. This assumption holds when $\Delta T$ is small. As $\Delta T$ increases, the underlying dynamics at $t + \Delta T$ would be different from the dynamics at $t$. Thus, the TANGO's performance degrades accordingly, and the advancement compared to RE-Net also vanishes.

\subsubsection{Inductive Link Prediction}
New graph nodes might emerge as time evolves in many real-world applications, i.e., new users and items. Thus, 
a good model requires a strong generalization power to 
deal with unseen nodes. We propose a new task, e.g., inductive link prediction, to validate the model potential in predicting the links regarding \textit{unseen entities} at a future time. A test quadruple is selected for the inductive prediction if either its subject or object or both haven't been observed in the training set. 
For example, in the test set of ICEWS05-15, we have a quadruple (\textit{Raheel Sharif}, \textit{express intent to meet or negotiate}, \textit{Chaudhry Nisar Ali Khan}, 2014-12-29). The entity \textit{Raheel Sharif} does not appear in the training set, indicating that the aforementioned quadruple contains an entity that the model does not observe in the training set. We call the evaluation of this kind of test quadruples the \textit{inductive link prediction analysis}. 

We perform the future link prediction on these inductive link prediction quadruples, and the results are shown in Table \ref{tab: inductive link prediction results}. We compare our model with the strongest baseline RE-Net on ICEWS05-15. We also report the results achieved by TANGO \includegraphics[height=0.8em]{TANGO.png} without the graph transition layer to show the performance boost brought by it. As shown in Table \ref{tab: inductive link prediction results}, TANGO-TuckER achieves the best results across all metrics. Both TANGO-TuckER and TANGO-Distmult can beat RE-Net, showing the strength of our model in inductive link prediction. The results achieved by the TANGO models are much better than their variants without the graph transition layers, which proves that the proposed graph transition layer plays an essential role in inductive link prediction.

\section{Conclusions}
We propose a novel representation method, TANGO\includegraphics[height=0.8em]{TANGO.png}, for forecasting future links on temporal knowledge graphs (tKGs). We propose a multi-relational graph convolutional layer to capture structural dependencies on tKGs and learn continuous dynamic representations using graph neural ordinary differential equations. Especially, our model is the first one to show that the neural ODE can be extended to modeling dynamic multi-relational graphs. Besides, we couple our model with the graph transition layer to explicitly capture the information provided by the edge formation and deletion. According to the experimental results, TANGO achieves state-of-the-art performance on five benchmark datasets for tKGs. We also propose two new tasks to evaluate the potential of link forecasting models, namely inductive link prediction and long horizontal link forecasting. TANGO performs well in both tasks and shows its great potential. 

\bibliography{anthology,custom}
\bibliographystyle{acl_natbib}

\appendix 
\section*{Appendix}
\begin{table*}[htbp]
    \centering
    \resizebox{\textwidth}{!}{
    \large\begin{tabular}{@{}|l|cccc|cccc|cccc|cccc|cccc|@{}}
        \hline
        Datasets & \multicolumn{4}{|c}{\textbf{ICEWS05-15 - raw}} &  \multicolumn{4}{|c}{\textbf{ICEWS14 - raw}} & \multicolumn{4}{|c|}{\textbf{ICEWS18 - raw}}& \multicolumn{4}{|c}{\textbf{WIKI - raw}}& \multicolumn{4}{|c|}{\textbf{YAGO - raw}}\\
\toprule
        Model & MRR & Hits@1 & Hits@3  & Hits@10 & MRR & Hits@1 & Hits@3  & Hits@10& MRR & HITS@1 & HITS@3 & HITS@10 & MRR & Hits@1 & Hits@3  & Hits@10 & MRR & Hits@1 & Hits@3  & Hits@10 \\
\midrule\relax
         Distmult & 24.55 & 15.85 & 27.53 & 42.17
         & 14.00 & 7.72 & 14.65 & 27.16 
         &  16.30 & 9.25 & 17.67 & 30.93 
         & 42.08 & 34.29 & 48.69 & 53.25 
         & 47.66 & 36.59 & 55.89 & 67.45\\
       TuckER & 26.95 & 16.81 & 29.69 & 47.61
       & 18.39 & 10.69 & 20.01 & 33.42 
       & 20.20 &  12.08 &  21.99 & 36.91 
       & \textbf{42.50} & \textbf{34.41} & \textbf{49.41} & 53.90 
       & 47.48 & 36.20 & 55.55 & 68.07
       \\
       COMPGCN & 29.41 & 20.41 & 32.17 & 47.65
       & 17.13 & 9.36 & 18.84 & 32.54 
       & 19.98 & 11.45 & 22.25 & 37.73 
       & 42.33 & 34.02 & 48.65 & \textbf{54.63} 
       & 47.08 & 65.36 & 66.90 & 68.81\\
\midrule
         TTransE & 20.89 & 4.88 & 3.11 & 49.66
         & 9.21 & 1.12 & 11.19 & 27.46 
         & 7.92 & 1.75 & 8.00 & 21.02 
         & 19.53 & 12.34 & 23.11 & 32.47 
         & 26.18 & 12.36 & 36.16 & 48.00\\
       TA-DistMult  & 24.03 & 14.37 & 27.36 & 44.04
       & 9.92 & 4.39 & 9.99 & 20.90 
       & 11.05 & 5.24 & 11.72 & 22.55 
       & 27.33 & 19.94 & 32.05 & 39.42 
       & 45.54 & 36.54 & 51.08 & 62.15\\

       RE-Net & 39.31 & 28.88 & 44.40 & 59.38
       & 23.84 & 14.60 & 26.48 & 42.58 
       & 26.62 & 16.91 & 30.26 & 45.82 
       & 31.10 & 25.31 & 34.13 & 41.33 
       & 46.28 & 37.52 & 51.77 & 61.55\\
\bottomrule
        \includegraphics[height=0.8em]{TANGO.png}-TuckER 
        & \textbf{41.82} & \textbf{31.10} & \textbf{47.55} & \textbf{62.19}
        & \textbf{24.36}  & \textbf{15.12} & \textbf{27.15} & \textbf{43.07} 
        & \textbf{27.59} & \textbf{17.77} & \textbf{31.40} & \textbf{46.92} 
        & 31.99 & 25.74 & 35.00 & 42.61 
        & 49.31 & 40.78 & 55.12 & 63.73\\
        \includegraphics[height=0.8em]{TANGO.png}-Distmult 
        & 40.23 & 30.53 & 44.95 & 59.05
        & 22.87  & 14.22 & 25.43 & 40.32
        & 26.21 & 16.92 & 29.77 & 44.41 
        & 32.53 & 26.33 & 35.75 & 43.17 
        & \textbf{49.49} & \textbf{40.90} & \textbf{55.42} & \textbf{63.74}\\
         \hline
    \end{tabular}}
    \caption{Future link prediction results on benchmark datasets. Evaluation metrics are raw MRR (\%) and Hits@1/3/10 (\%). \includegraphics[height=0.8em]{TANGO.png} denotes TANGO. 
    The best results are marked in bold.}\label{tab: raw link prediction results}
\end{table*}

\begin{table*}[t]
    \centering
    \resizebox{\textwidth}{!}{
    \large\begin{tabular}{|l|cccc|cccc|cccc|cccc|cccc|}
        \hline
        Datasets & \multicolumn{4}{|c}{\textbf{ICEWS05-15 - unaware filtered}} &  \multicolumn{4}{|c}{\textbf{ICEWS14 - unaware filtered}} & \multicolumn{4}{|c}{\textbf{ICEWS18 - unaware filtered}} & \multicolumn{4}{|c}{\textbf{WIKI - unaware filtered}}& \multicolumn{4}{|c|}{\textbf{YAGO - unaware filtered}}\\
\toprule
        Model & MRR & Hits@1 & Hits@3  & Hits@10 & MRR & Hits@1 & Hits@3  & Hits@10 & MRR & Hits@1 & Hits@3 & Hits@10 & MRR & Hits@1 & Hits@3  & Hits@10 & MRR & Hits@1 & Hits@3  & Hits@10\\
\midrule\relax
         Distmult & 48.77 & 43.85 & 51.22 & 57.99 
         & 33.88 & 27.86 & 36.16 & 45.14 
         &  40.28 & 36.04 & 41.78 & 48.36 
         & 53.22 & 52.61 & 53.41 & 54.20 
         & 67.55 & 66.76 & 67.49 & 69.11\\
       TuckER & 58.69 & 54.74 & 59.82 & 66.57
       & 46.51 & 41.11 & 49.45 & 57.34 
       & 44.50 &  38.33 &  46.11 & 53.71 
       & 53.97 & \textbf{52.70} & \textbf{54.15} & 54.94 
       & 67.40 & 66.22 & 67.62 & 69.84
       \\
       COMPGCN & 49.60 & 43.13 & 52.85 & 61.59 &
       38.15 & 31.04 & 41.00 & 51.44 &
        35.68 & 27.87 & 39.38 & 49.94 &
        53.54 & 52.29 & 53.61 & 55.76 &
        66.66 & 65.36 & 66.90 & 68.81\\
\midrule
         TTransE & 28.81 & 5.83 & 48.67 & 60.38
         & 15.95 & 1.57 & 25.98 & 42.67 
         & 10.52 & 3.01 & 11.98 & 26.16 
         & 31.94 & 24.82 & 36.91 & 43.55 
         & 33.73 & 20.99 & 43.51 & 52.61\\
       TA-DistMult & 38.54 & 29.94 & 42.92 & 54.81 
       & 18.74 & 11.97 & 20.32 & 31.95 
       & 16.27 & 10.22 & 17.39 & 27.91 
       & 50.18 & 48.65 & 51.41 & 52.37 
       & 66.06 & 64.36 & 66.78 & 68.74\\
       RE-Net & 57.66 & 51.86 & 60.40 & 68.60 
       & 45.24 & 37.82 & 48.53 & 58.92 
       & 43.02 & 36.26 & 45.61 & 56.03 
       & 52.27 & 50.92 & 52.73 & 53.57 
       & 64.68 & 62.94 & 65.11 & 67.82\\
\bottomrule
        \includegraphics[height=0.8em]{TANGO.png}-TuckER
        & \textbf{59.93} & \textbf{54.99} & \textbf{62.65} & \textbf{69.64} &
        46.42  & 38.94 & \textbf{50.25} & \textbf{59.80} 
        & \textbf{44.56} & 37.87 & \textbf{47.46} & \textbf{57.06} & 
        53.28 & 52.21 & 53.61 & 54.84 &
        67.21 & 65.56 & 67.59 & 70.04\\
        \includegraphics[height=0.8em]{TANGO.png}-Distmult 
        & 58.89 & 54.42 & 60.76 & 67.47& \textbf{46.68}  
        & \textbf{41.20} & 48.64 & 57.05
        & 44.00 & \textbf{38.64} & 45.78 & 54.27  & 
        \textbf{54.05} & 51.52 & 53.84 & \textbf{55.46} &
        \textbf{68.34} & \textbf{67.05} & \textbf{68.39} & \textbf{70.70}\\
         \hline
    \end{tabular}}
    \caption{Future link prediction results on benchmark datasets. Evaluation metrics are time-unaware filtered MRR (\%) and Hits@1/3/10 (\%). \includegraphics[height=0.8em]{TANGO.png} denotes TANGO. 
    The best results are marked in bold.}\label{tab: tid-filtered link prediction results}
\end{table*}


\begin{table*}[t]
    \centering
    \resizebox{\textwidth}{!}{
    \large\begin{tabular}{|l|cccc|cccc|cccc|cccc|cccc|}
        \hline
        Datasets & \multicolumn{4}{|c}{\textbf{ICEWS05-15 - aware filtered}} & \multicolumn{4}{|c}{\textbf{ICEWS18 - aware filtered}} & \multicolumn{4}{|c}{\textbf{WIKI - aware filtered}}& \multicolumn{4}{|c|}{\textbf{YAGO - aware filtered}}\\
\toprule
        Model & MRR & Hits@1 & Hits@3  & Hits@10 & MRR & Hits@1 & Hits@3 & Hits@10 & MRR & Hits@1 & Hits@3  & Hits@10 & MRR & Hits@1 & Hits@3  & Hits@10\\
\midrule\relax

        \includegraphics[height=0.8em]{TANGO.png}-TuckER
        & 44.57 & 34.40 & 49.94 & 63.95
        & 30.68 & 20.75 & 34.61 & 50.43 & 
        62.29 & 59.54 & 63.92 & 66.63 &
        69.29 & 64.33 & 72.40 & 77.63\\
        \includegraphics[height=0.8em]{TANGO.png}-Distmult 
        & 43.33 & 33.46 & 48.45 & 62.05
        & 29.62 & 20.18 & 33.35 & 48.36  &
        63.93 & 62.14 & 64.74 & 67.06 &
        70.79 & 66.15 & 74.04 & 78.18\\
         \hline
    \end{tabular}}
    \caption{Validation results on benchmark datasets regarding our model. Evaluation metrics are time-aware filtered MRR (\%) and Hits@1/3/10 (\%). \includegraphics[height=0.8em]{TANGO.png} denotes TANGO. 
    The best results are marked in bold. ICEWS14 has no validation set.}\label{tab: td-filtered validation results}
\end{table*}

\section{Representation Inference}
\label{app: representation inference}
\begin{figure}[htbp]
    \centering
    \includegraphics[scale=0.45]{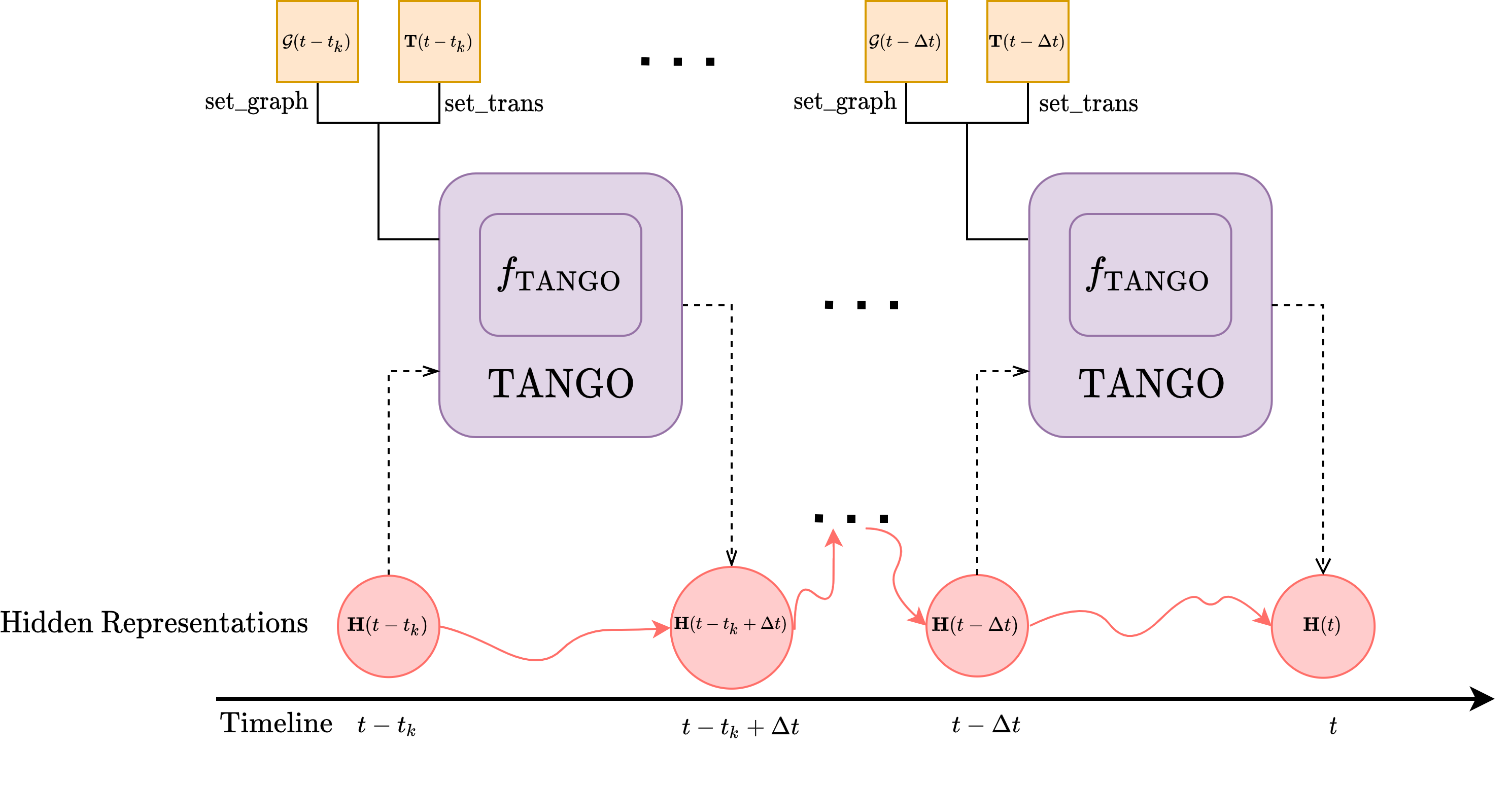}
    \caption{Illustration of the inference procedure. The shaded purple area represents the whole architecture of TANGO. It is a Neural ODE equipped with a GNN-based module $f_{\textrm{TANGO}}$. Dashed arrows denote the input and the output path of the graph's hidden state. Red solid arrows indicate the continuous hidden state flows learned by TANGO. Black solid lines represent that TANGO calls the function $\textrm{set\_graph}$ and $\textrm{set\_trans}$. The corresponding graph snapshots $\mathcal{G}$ and transition tensors $\textbf{T}$ are input into $f_{\textrm{TANGO}}$ for learning temporal dynamics.}
    \label{fig:TANGOforward}
\end{figure}

Assume we want to forecast a link at $t$. We take the graph histories between the timestamp $(t-t_k)$ and the timestamp $t$ into account, where $t_k$ indicates the length of history. To infer the hidden representations $\mathbf H(t)$, we first use the initial embeddings $\textrm{Emb}(\mathcal V, \mathcal R)$ to approximate  the hidden representations $\textbf{H}(t-t_k)$.
Then we take $\textbf{H}(t-t_k)$ as the NODE input at the timestamp $(t-t_k)$, and integrate it with an ODE solver $\textrm{ODESolver}(\textbf{H}(t-t_k), f_{\textrm{TANGO}}, t-t_k, t, \Theta_{\textrm{TANGO}}, \mathcal G)$ over time. 
As the hidden state evolves with time, it learns from different graph observations taken at different time. The whole process is described in Figure \ref{fig:TANGOforward} and Algorithm 1. In Figure \ref{fig:TANGOforward}, $\textrm{set\_graph}$ and $\textrm{set\_transition}$ stand for two functions used to feed graph snapshots and the transition tensors into the neural network $f_{\textrm{TANGO}}$. They are called at every observation time before integration. 

\section{Evaluation Metrics}
\label{app: evaluation metrics}
We report the results in three settings, namely raw, time-unaware filtered, and time-aware filtered. For time-unaware filtered results, we follow the filtered evaluation constraint applied in \cite{10.5555/2999792.2999923, jin2019recurrent},  where we remove from the list of corrupted triplets all the triplets that appear either in the training, validation, or test set except the triplet of interest. Time-unaware filtering setting is inappropriate for temporal KG reasoning, while  the time-aware filtering setting provides fairer results. For time-aware filtered results, we follow the setting proposed by \cite{han2021xerte} by only removing from the list of corrupted triplets all the triplets that appear at the query time $t_q$. 
The following example illustrates the reason why the time-aware filtered results are fairer than the time-unaware filtered results. Assume we have a test quadruple of interest (\textit{Xi Jinping}, \textit{make a visit}, \textit{New Zealand}, 2014-11-26) in the test set, and we derive an object prediction query (\textit{Xi Jinping}, \textit{make a visit}, $?$, 2014-11-26) from this quadruple where the query time is 2014-11-26. Additionally, we have another quadruple (\textit{Xi Jinping}, \textit{make a visit}, \textit{South Korea}, 2014-07-05) in the test set. According to the time-unaware filtering setting \cite{10.5555/2999792.2999923}, (\textit{Xi Jinping}, \textit{make a visit}, \textit{South Korea}) will be filtered out since it appears in the test set. However, it is unreasonable because (\textit{Xi Jinping}, \textit{make a visit}, \textit{South Korea}) is not valid at 2014-11-26. Therefore, we use the time-aware filtered setting, which, in our example, will only filter the triplets (\textit{Xi Jinping}, \textit{make a visit}, $o$) appearing at 2014-11-26. Here, $o$ denotes all the objects from triplets accompanied with \textit{Xi Jinping}, \textit{Make a visit}, and the date 2014-11-26.  

\section{Implementation Details}
\label{app: Implementation Details}
We train TANGO with the following settings. We tune the model across a range of hyperparameters as shown in Table \ref{tab: hyperparameter search space}. We do 432 trials, and each trial runs 20 epochs. We select the best-performing configuration according to filtered MRR on validation data. The best configuration will be further trained until its convergence. We run the selected configuration five times and obtain an averaged results. Specifically, 
we use a fixed-grid ODE solver, fourth-order Runge-Kutta, as the ODE solver, and implement the interpolated reverse dynamic method \cite{daulbaev2020interpolated} with 3 Chebyshev nodes to keep training time tractable while maintaining high precision.  To improve the ODE solver's precision, we re-scale the time range of each dataset from 0 to 0.01 (or 0.1). This step restricts the length of ODE integration, preventing the high error induced by ODE solvers. For each query, we set the time range of the input history $t_k$ to 4 days for the ICEWS datasets. For WIKI and YAGO, we set $t_k$ to 4 years. Besides, we choose different values for the transition coefficient $w$ for different datasets. Our model is implemented with PyTorch \cite{paszke2019pytorch}, and the experiments are run on GeForce RTX 2080 Ti. A detailed report of the best configuration is provided in Table \ref{tab: Best hyperparameter setting}. 

We implement Distmult in PyTorch and use the binary cross-entropy loss for learning parameters. We use the official implementation of $\,$TuckER\footnote{https://github.com/ibalazevic/TuckER}, 
 COMPGCN\footnote{https://github.com/malllabiisc/CompGCN}, and RE-Net\footnote{https://github.com/INK-USC/RE-Net}. For a fair comparison, we choose to use the variant of RE-Net with ground truth history during multi-step inference, and thus the model knows all the interactions before the time for testing. Besides, we set the history length of RE-Net to 10 and use the max-pooling in the global model. 
Additionally, we use the implementation of TTransE and TA-Distmult provided in \cite{jin2019recurrent}. For TA-Distmult, the vocabulary of temporal tokens consists of year, month, and day for all the datasets. 
We use the released code to implement DE-SimplE\footnote{https://github.com/BorealisAI/de-simple}, TNTComplEx\footnote{https://github.com/facebookresearch/tkbc}, and CyGNet\footnote{https://github.com/CunchaoZ/CyGNet}. 
All the baselines are trained with Adam Optimizer \cite{kingma2017adam}, and the batch size is set to 512.
 
 
 \begin{table}[ht!]
\caption{Search space of hyperparameters. $w$ represents the weight controlling how much the model learns from edge formation and dissolution. \textit{Scale} represents the time range re-scaling parameter as introduced in \ref{app: Implementation Details}.}
\begin{center}
\resizebox{.7\columnwidth}{!}{
    \begin{tabular}{ll}
    \hline
    Hyperparameter                             & Search space                          \\ \hline
    Embedding size                        & \{200, 300\}     \\
    \# MGCN layer                        & \{2, 3\}     \\
    Decoder                        & \{TuckER, Distmult\}     \\
    \textit{Scale}                        & \{0.001, 0.01, 0.1\}     \\
    $w$                               & \{0.01, 0.1, 1\}     \\
    Dropout                               & \{0.3, 0.5\}     \\
    History length                      & \{4, 6, 10\}      \\
    \hline
    \end{tabular}
    }
\end{center}
\label{tab: hyperparameter search space}
\end{table}

\begin{table}[ht!] 
    \caption{Best hyperparameter settings on each dataset.}
    \label{tab: Best hyperparameter setting}
    \begin{center}
      \resizebox{\columnwidth}{!}{
    \begin{tabular}{lcccccc} 
      \toprule 
     \multicolumn{1}{l}{Datasets} & \multicolumn{1}{c}{\textbf{ICEWS14}} & \multicolumn{1}{c}{\textbf{ICEWS18}} &\multicolumn{1}{c}{\textbf{ICEWS05-15}} & \multicolumn{1}{c}{\textbf{WIKI}}
     & \multicolumn{1}{c}{\textbf{YAGO}}\\
      \midrule 
      Hyperparameter &   &  &  &    & \\
       \midrule 
       Embedding size & 200     &  200&  200&  200& 300\\
       \# MGCN layer &  2   &  2&   2&   2& 3\\
       Decoder &   TuckER &  TuckER&  TuckER&  Distmult& Distmult\\
       Scale &  0.01&  0.1&  0.1&   0.1& 0.1\\
       $w$ &   0.01&  1&   0.01&   1& 1\\
       Dropout &  0.3&  0.3&  0.3&  0.3& 0.3\\
       History length & 4& 4& 4&  4& 4\\
       
      \bottomrule 
    \end{tabular} }
    \end{center}
\end{table}	
 
 \section{Datasets}
 \label{app: dataset statistics}
 Table \ref{tab: data}
We follow the data preprocessing method and the dataset split strategy proposed in \cite{jin2019recurrent}. Specifically, we split each dataset except ICEWS14 in chronological order into three parts, e.g., $80\%/10\%/10\%$ (training/validation/test). For ICEWS14, we split it into the training set and testing set with $50\%/50\%$ since ICEWS14 is not provided with a validation set.  As explained in \cite{jin2019recurrent}, the difference between the first type (ICEWS) and the second type (WIKI and YAGO) of tKG datasets is that the first type datasets are events that often last shortly and happen multiple times, i.e., Obama visited Japan four times. In contrast, the events in the second type datasets last much longer and do not occur periodically, i.e., Eliran Danin played for Beitar Jerusalem FC between 2003 and 2010.
\begin{table}[htbp]
    \centering
    \resizebox{\columnwidth}{!}{
\begin{tabular}{c c c c c c c c} \hline
Dataset&$N_\textrm{train}$&$N_\textrm{valid}$&$N_\textrm{test}$&$|\mathcal{V}|$&$|\mathcal R|$&$N_{\textrm{obs}}$\\ \hline
ICEWS14 \cite{trivedi2017knowevolve} & $323,895$ & $-$ & $341,409$ & $12,498$ & $260$ & $365$\\ 
ICEWS18 \cite{DVN/28075_2015} & $373,018$ & $45,995$ & $49,545$ & $23,033$ & $256$ & $304$\\ 
ICEWS05-15 \cite{garcia-duran-etal-2018-learning} & $369,104$ & $46,188$ & $46,037$ & $10,488$ & $251$ & $4,017$\\ 
WIKI \cite{leblay2018deriving} & $539,286$ & $67,538$ & $63,110$ & $12,554$ & $24$&$232$\\ 
YAGO \cite{mahdisoltani2013yago3} & $161,540$ & $19,523$ & $20,026$ & $10,623$ & $10$ & $189$\\ \hline
\hline
\end{tabular}}
\caption{Dataset statistics. $N_\textrm{train}$, $N_\textrm{valid}$, $N_\textrm{test}$ represent the number of quadruples in the training set, validation set, and test set, respectively. $N_{\textrm{obs}}$ denotes the number of observations, where we take a snapshot of the tKG at each observation.}
\label{tab: data}
\end{table}

\section{Impact of Past History Length}
\begin{figure}[htbp]
    \begin{center}
    \begin{tikzpicture}[scale=0.5, transform shape]
        \pgfplotsset{width=10cm, height=8cm, compat=1.16, legend style={at={(0.55,0.15)},anchor=west}}
      \definecolor{clr1}{HTML}{12AAB5}
      \definecolor{clr2}{HTML}{E85642}
    \begin{axis}[%
      axis y line*=left,
      xlabel=History Length,
      ylabel=Time-aware Filtered MRR,
      xtick={1,2,3,4,5,6,7},
      ytick={39,40,41,42,43},
      ymin=39, 
      xmin=1.0,xmax=7, 
    ]
      \addplot+[mark=square*, mark options={fill=clr2}, clr2] coordinates 
      {(1,39.823) (2,40.012) (3, 41.186) (4, 42.858) (5, 42.953) (6, 43.163) (7, 43.277)};
      \label{fig:history}
    
    \end{axis}
    \begin{axis}[%
        axis x line=none,
        axis y line*=right,
        ylabel=Training Time/ Epoch (s),
        xtick={1,2,3,4,5,6,7},
        xtick distance={50},
        ytick={0,500,1000,1500,2000,2500},
        ymin=0,ymax=3000,
        xmin=1.0,xmax=7 
      ]
        \addlegendimage{/pgfplots/refstyle=fig:history}\addlegendentry{MRR}
        \addplot+[mark=square*, mark options={fill=clr1}, clr1] coordinates 
        {(1,304) (2,708) (3, 1029) (4, 1326) (5, 1640) (6, 1973) (7, 2265)};
        \addlegendentry{Training Time}
      
    \end{axis}
    \end{tikzpicture}
    \end{center}
    \caption{Time-aware filtered MRR (\%) and Training Time (seconds) on ICEWS05-15 corresponding to different history length (days).}
    \label{fig: length of history and run time}
\end{figure}
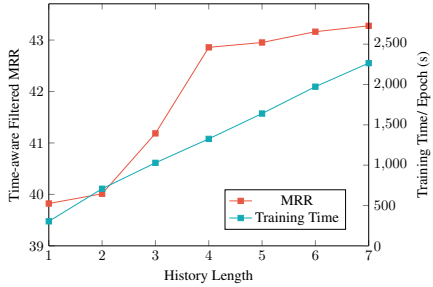

As mentioned in \ref{app: representation inference}, TANGO utilizes the previous histories between $(t - t_k)$ and $t$ to forecast a link at $t$, where $t_k$ is a hyperparameter. Figure \ref{fig: length of history and run time} shows the
performance with various lengths of past histories along with the corresponding training time. When TANGO uses longer histories, MRR is getting higher. However, a long history requires more forwarding inferences. The choice of history length is a trade-off between the performance and computational cost. We observe that the gain of MRR compared to the training time is not significant when the length of history is four and over. Thus, the history length of four is chosen in our experiments.

\begin{figure}[htbp]
\centering
\includegraphics[width=.7\columnwidth]{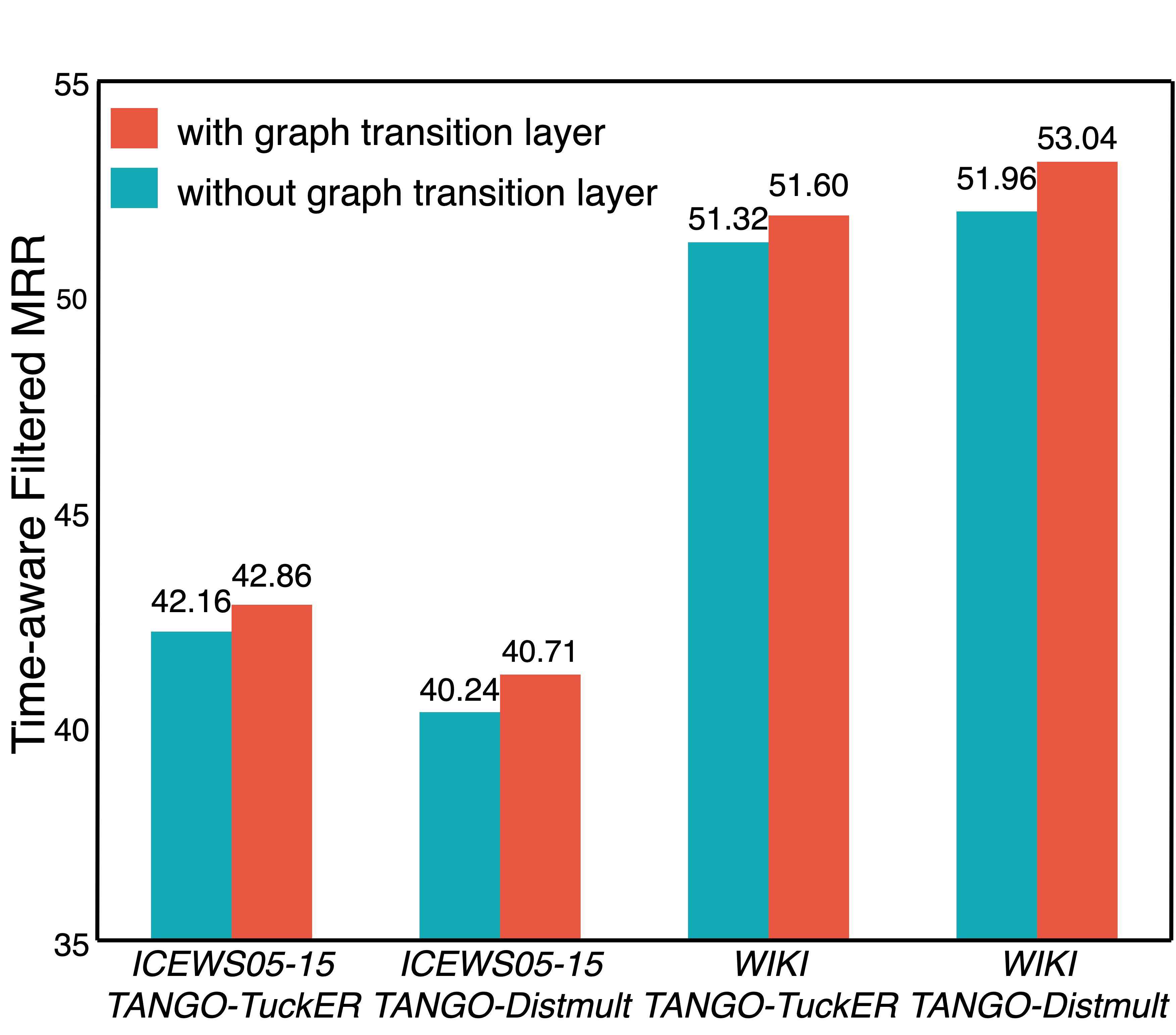}
\caption{\label{fig: overall ablation.} Time-aware filtered MRR of TANGO with or without  the graph transition layer  on  the whole test sets  of ICEWS05-15 and WIKI.}
\end{figure}

\begin{table}[t]
    \centering
    \resizebox{\columnwidth}{!}{
    \begin{tabular}{|l|cccc|cccc|}
        \hline
        Datasets & \multicolumn{4}{|c|}{\textbf{ICEWS05-15 continuous - aware filtered}}
        & \multicolumn{4}{|c|}{\textbf{ICEWS05-15 - aware filtered}}\\
\toprule
        Model & MRR & Hits@1 & Hits@3  & Hits@10 & MRR & Hits@1 & Hits@3  & Hits@10\\
\midrule\relax
        TTransE 
        & 20.55 & 5.36 & 29.80 & 47.54
        & 21.24 & 4.98 & 31.48 & 49.88\\
        CyGNet
        & 34.13 & 25.06 & 37.85 & 51.94
        & 35.79 & 26.09 & 40.18 & 54.48\\
        DE-SimplE
        & 33.56 & 24.79 & 37.32 & 50.63
        & 35.57 & 26.33 & 39.41 & 53.97\\
        TNTComplEx 
        & 33.96 & 24.93 & 37.86 & 51.30
        & 35.88 & 26.92 & 39.55 & 53.43\\
        \includegraphics[height=0.8em]{TANGO.png}-TuckER
        & \textbf{37.69} & \textbf{28.01} & \textbf{45.00} & \textbf{59.05}
        & \textbf{42.86} & \textbf{32.72} & \textbf{48.14} & \textbf{62.34}\\
        \includegraphics[height=0.8em]{TANGO.png}-Distmult 
        & 36.91 & 26.91 & 40.28 & 54.34
        & 40.71 & 31.23 & 45.33 & 58.95\\
         \hline
    \end{tabular}}
    \caption{Future link prediction results on ICEWS05-15 continuous dataset. Evaluation metrics are time-aware filtered MRR (\%) and Hits@1/3/10 (\%). \includegraphics[height=0.8em]{TANGO.png} denotes TANGO. 
    The best results are marked in bold.}\label{tab: td-filtered continuous results}
\end{table}

\begin{table}[htbp]
    \centering
    \resizebox{\columnwidth}{!}{
\begin{tabular}{c c c c c c c c} \hline
Dataset&$N_\textrm{train}$&$N_\textrm{valid}$&$N_\textrm{test}$&$|\mathcal{V}|$&$|\mathcal R|$&$N_{\textrm{obs}}$\\ \hline
ICEWS05-15 continuous & $149,001$ & $17,962$ & $17,902$ & $10,488$ & $251$ & $1,589$\\

\hline
\end{tabular}}
\caption{Dataset statistics. $N_\textrm{train}$, $N_\textrm{valid}$, $N_\textrm{test}$ represent the number of quadruples in the training set, validation set, and test set, respectively. $N_{\textrm{obs}}$ denotes the number of observations, where we take a snapshot of the tKG at each observation.}
\label{tab: continuous data}
\end{table}

\section{Analysis on Temporal KGs with Irregular Time Intervals}
Most existing tKG reasoning models cannot properly deal with temporal KGs with irregular time intervals, while TANGO model them much better due to the nature of Neural ODE. We verify this via experiments on a new dataset. We call it \textit{ICEWS05-15\_continuous}. We sample the timestamps in ICEWS05-15 and keep the time intervals between each two of them in a range from 1 to 4. We only keep the temporal KG snapshots at the sampled time and extract a new subset. 
ICEWS05-15\_continuous fits the setting when observations are taken non-periodically in continuous time. The dataset statistics of  ICEWS05-15\_continuous is reported in Table \ref{tab: continuous data}. We train our model and baseline methods on it and evaluate them with time-aware filtered MRR. As shown in Table \ref{tab: td-filtered continuous results}, we validate that TANGO performs well on temporal KGs with irregular time intervals.

\section{Average runtime for each approach}
Table  \ref{tab: Average runtime and number of parameters for each approach} show  the average runtime for each model.

\begin{table}[ht!] 
    \caption{Average training time (second) until convergence}
    \label{tab: Average runtime and number of parameters for each approach}
    \begin{center}
      \resizebox{\columnwidth}{!}{
    \begin{tabular}{lcccccc} 
      \toprule 
     \multicolumn{1}{l}{Datasets} & \multicolumn{1}{c}{\textbf{ICEWS14}} & \multicolumn{1}{c}{\textbf{ICEWS18}} &\multicolumn{1}{c}{\textbf{ICEWS05-15}} & \multicolumn{1}{c}{\textbf{WIKI}}
     & \multicolumn{1}{c}{\textbf{YAGO}}\\
      \midrule 
      Model &  Runtime & Runtime & Runtime  & Runtime & Runtime\\
      \midrule 
      Distmult & 743     &  1,365&  401& 2,245& 3,310\\
      TuckER &  730   &  3,147&   1,626& 5,093& 2,795\\
      COMPGCN &   9,226 &  6,432&  1,607& 5,810& 2,233\\
      TTransE &  15,840&  23,894&  35,520&   19,337& 5,395\\
      TA-Distmult &   6,232&  112,188&   110,460&   83,999& 27,833\\
      RE-Net &  33,313&  46,068&  190,076&  42,983& 27,489\\
      \includegraphics[height=0.8em]{TANGO.png}-TuckER &5,796& 3,786& 15,301&  9,218& 2,355\\
      \includegraphics[height=0.8em]{TANGO.png}-Distmult &3,593& 2,883& 11,085&  15,086& 5,106\\
      \bottomrule 
    \end{tabular} }
    \end{center}
\end{table}	
\end{document}